\def\year{2019}\relax
\documentclass[letterpaper]{article} 
\usepackage{aaai19}  
\usepackage{times}  
\usepackage{helvet}  
\usepackage{courier}  
\usepackage{url}  
\usepackage{graphicx}  
\usepackage{amssymb}
\usepackage{amsmath}
\usepackage[sort&compress,square,comma,authoryear]{natbib}

\usepackage{amsfonts}
\usepackage{amssymb}
\usepackage{amsthm}
\usepackage{bbm}
\usepackage{latexsym}
\usepackage{mathtools}
\usepackage{color}

\usepackage{algorithm}
\usepackage{algorithmic}

\DeclareMathOperator*{\argmax}{argmax}

\newcommand{\km}[1]{{\color{black} #1}} 

\begin{document}
%
\title{Facility Locations Utility for Uncovering Classifier Overconfidence}
\author{Karsten Maurer$^1$ \hspace{.2in} Walter Bennette$^2$\\
$^1$Miami University Department of Statistics - maurerkt@miamioh.edu\\ 
$^2$Air Force Research Lab Information Directorate - walter.bennette.1@us.af.mil \\
}

\maketitle
\begin{abstract}
Assessing the predictive accuracy of black box classifiers is challenging in the absence of labeled test datasets. In these scenarios we may need to rely on a human oracle to evaluate individual predictions; presenting the challenge to create query algorithms to guide the search for points that provide the most information about the classifier's predictive characteristics. Previous works have focused on developing utility models and query algorithms for discovering unknown unknowns --- misclassifications with a predictive confidence above some arbitrary threshold. However, if misclassifications occur at the rate reflected by the confidence values, then these search methods reveal nothing more than a proper assessment of predictive certainty. We are unable to properly mitigate the risks associated with model deficiency when the model's confidence in prediction exceeds the actual model accuracy. We propose a facility locations utility model and corresponding greedy query algorithm that instead searches for overconfident unknown unknowns. Through robust empirical experiments we demonstrate that the greedy query algorithm with the facility locations utility model consistently results in oracle queries with superior performance in discovering overconfident unknown unknowns than previous methods. 
\end{abstract}

\section{Introduction}

Techniques such as active learning \citep{Settles2010} and domain adaptation \citep{Patel2014} can be used to create machine learning classifiers when large labeled datasets are not available for a specific task.  For example, the black box classifiers made available through many online services (Google Cloud, Amazon Web Services, etc.) require no training data and can be thought of as a kind of domain adaptation.  However, with limited amounts of labeled data, users may not be able to properly evaluate a model, and are left hoping the model will be useful for their intended task.  In this paper we build upon previous work to develop a human-in-the-loop method to help evaluate classifiers in the absence of labeled data.  Specifically, we develop an interactive method to uncover \textit{unknown unknowns} (UUs): points where a classifier is confident in its prediction, but wrong \citep{Attenberg2015}.  

Intuitive methods can be used to evaluate the performance of a model in the absence of labeled data.  For example, given a labeling budget one could sample points following an experimental design, sample points with the lowest classifier confidence, or sample points identified as informative to the classifier through active learning strategies.  These methods could provide a sense of a model's performance but will potentially miss high confidence mistakes. In cases where the correct classification of a specific class is critical for subsequent decision making (medical diagnoses, cyber security, forensic science), UUs can be defined with respect to misclassifications of the critical class.

UUs can be thought of as blind spots to a classification model, and can be caused by dataset bias during training \citep{stock2017convnets}, domain shift during use \citep{sugiyama2017dataset}, lack of model expressibility, and other causes of poor model fit. \citet{Lakkaraju2016} describe a classifier trained on a biased image dataset of cats with light fur and dogs with dark fur. When this classifier is used for inference it predicts that dogs with light fur belong to the cat class with high confidence.  The light fur dogs are UUs for the classifier and reveal a deficiency of the model.  

From the viewpoint of a rational actor, UUs represent costly mistakes because minimal risk mitigation strategies will have been deployed for these high confidence predictions.  The discovery of UUs could then allow new mitigation strategies to be formulated \citep{Nushi2016a}. Additionally, as further enumerated in \citet{Bansal2018}, finding UUs is valuable to understand classifier strengths and weaknesses and possibly avoid certain adversarial attacks.

Previous works have focused on discovering UUs defined to be misclassifications with a predictive confidence above some arbitrary threshold, $\tau$ (typically set to $0.65$ for binary classification).  With this definition it should be expected that $(1-\tau)$\% of points sampled at the threshold will be called an UU. This definition ignores the uncertainties of predictive modeling and the purpose of confidence scores.  We believe it is more valuable to uncover misclassifications where the rate of UU discovery is higher than should be expected based on predictive confidence, thus searching for classifier \textit{overconfidence}. Otherwise, the discovered UUs may reveal nothing more than the fact that the model is performing as expected.  Alternatively, uncovering overly confident UUs can reveal problematic areas of the classifier and begin to hint at mitigation strategies such as model calibration \citep{bella2010calibration}.  

In the following manuscript we first discuss the established algorithms for discovering UUs, and demonstrate deficiencies in the utility design of previous methods. We then propose our own facility locations utility model and corresponding search algorithm. Through robust empirical experiments we demonstrate that the greedy query algorithm with the facility locations utility model consistently results in oracle queries with superior performance in discovering overconfident UUs than previous methods. We conclude with a discussion of these results, access to the implementation and avenues for future work.

\section{Previous Works}

The search for unknown unknowns of a classification model operates with an unlabeled test set and does not require access to the original training features. This type of scenario can arise, for example, with an externally provided black box classifier.  It is also assumed that an oracle can be queried to provide labels up to a certain budget and that the model can provide a realistic confidence of its prediction. Given these assumptions the search for UUs is carried out over a set of unlabeled points for which a classifier has provided predicted labels and associated confidence values.  

\citet{Attenberg2015} turned the search for classifier errors into a game to be played by humans called "Beat the Machine". The intent of the game was to discover websites that would fool a hate-speech classifier. Through trial and error, a utility function was derived to value high confidence mistakes (UUs) from diverse URLs. Then users ``played'' the game by submitting URLs to earn a monetary reward tied to the utility function. 

Semi-automated methods to search for UUs have also been proposed [\citealt{Lakkaraju2016}; \citealt{Bansal2018}].  Each of these methods can be distilled to the same basic components.  First, a utility function is constructed to capture the value of a set of discovered UUs. Second, a strategy is developed to sample unlabeled points to maximize the designed utility, where each search strategy is driven by some estimation of a point's likelihood of being an UU.  Third, all methods execute a search following the developed strategy until a labeling budget is exhausted.

\citet{Lakkaraju2016} introduced the first algorithmic approach for discovering UUs with a semi-automated search directly providing unlabeled points to an oracle.  Their utility function provides a unit value for each discovered UU and penalizes by the cost of labeling (for example the number of words read to evaluate a text classification). Their search strategy relies on a multi-armed bandits approach to sample from clusters of the points based on classifier confidence and a derived feature space.  The feature space is not restricted to match that of the classifier to accommodate cases when the original features are unavailable.  The bandit search is driven by tracking the average utility of a cluster, which can be viewed as an indication of the likelihood of finding an UU in that cluster. 

\citet{Bansal2018} argue that the unit utility of \citet{Lakkaraju2016} motivates the discovery of very similar UUs. Instead, they propose an adaptive coverage-based utility model that attempts to encourage the discovery of high confidence UUs spread throughout a feature space. They then search for UUs via a greedy algorithm to maximize utility.  Like the bandit search, the greedy search relies on a clustering of a derived feature space and is driven by the observed ratio of UUs in each cluster. 

\begin{figure*}[h!]
  \centering
  \includegraphics[width=\textwidth]{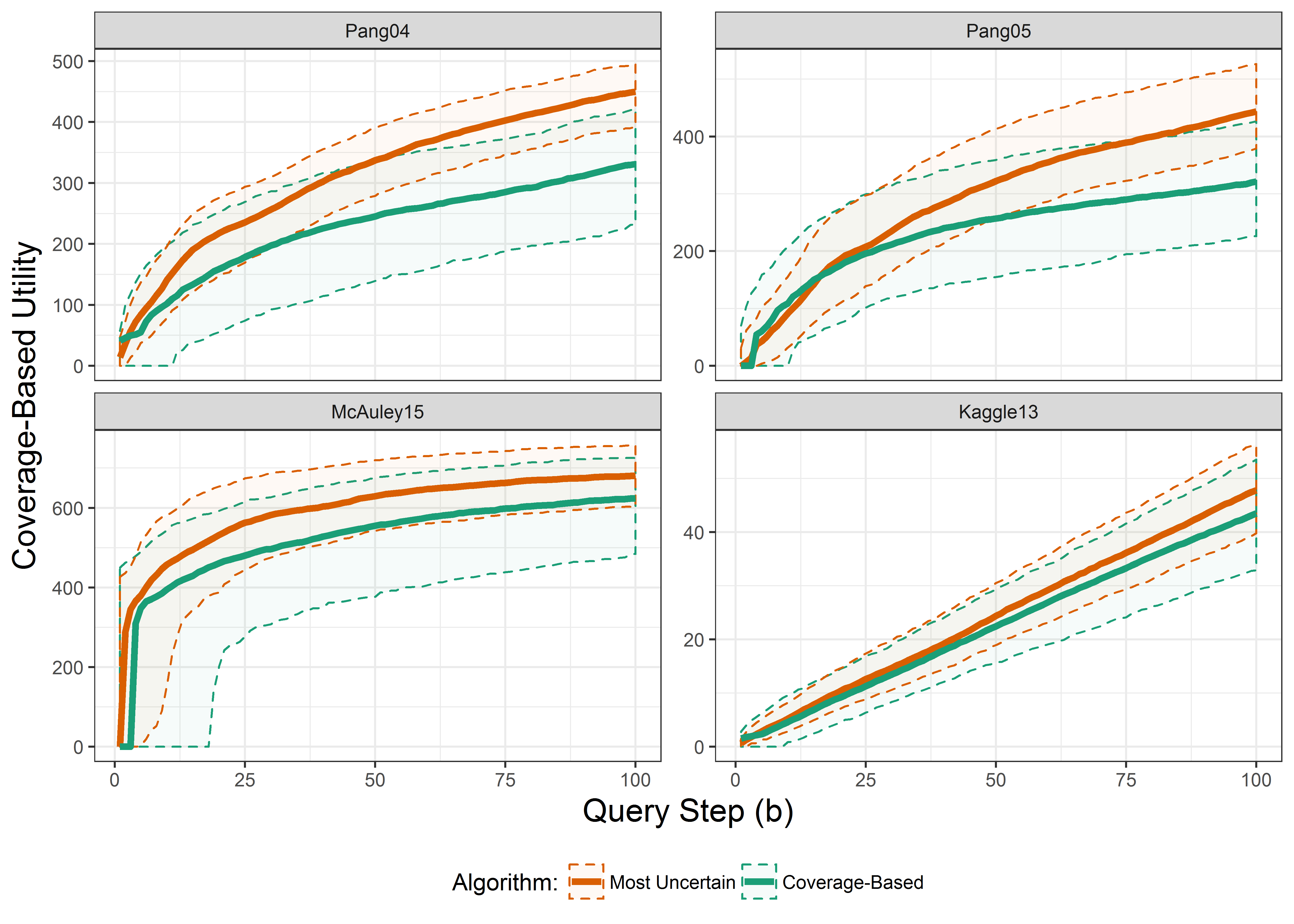}
  \caption{Comparison of coverage-based utility outcomes achieved under most uncertain and coverage-based search algorithms. Monte Carlo medians (solid) and 90\% prediction bands (dashed).  In all datasets, higher coverage-based utility is achieved with most uncertain searches.}
  \label{fig:coverutil}
\end{figure*}

Looking closer at the coverage-based utility model, it sums the prediction confidence of every test point multiplied by a similarity measure comparing it to its closest discovered UU.  It has the form: $$U(Q) = \sum_{x \in \mathbb{X}} c_x \cdot \max_{q \in S} \left\{sim\left(x,q \right) \right\}$$ where $\mathbb{X} \subset \mathbb{R}^p$ is the set of available $p$-dimensional unlabeled test points, $Q \subset \mathbb{X}$ is the set of points labeled by an oracle, $S = \left\{x|x \in Q, y_x \neq M(x)\right\}$ is the set of discovered UUs for some classifier $M(x):\mathbb{X} \rightarrow class$, $c_x$ is the classifier's confidence in its prediction of $x$, and $sim(x,q)$ is a distance-based similarity metric. 

Given this utility model the search for UUs is performed by greedily selecting the point $q'$ that maximizes the expected utility increase.  Meaning, $q'$ is selected to maximize, $$E\left[U_x\left(Q \cup q'\right)\right] = \hat{\phi}(x) \cdot c_x \cdot \max_{q \in S \cup q'} \left\{sim\left(x,q \right) \right\},$$ where $\hat{\phi}(x) = P\left(y_x \neq M(x) |Q \right)$ is the cluster conditional probability that $x$ is misclassified given the query set.  As previously stated, this method is designed to incentivize a broader search for UUs and gives higher utility for finding misclassifications in higher confidence regions.  

Unfortunately, the coverage-based utility search consistently achieves lower utility than the simple strategy of sequentially querying points for which the classifier is most uncertain.  This is shown in Figure~\ref{fig:coverutil} which displays Monte Carlo medians and 90\% predictions bands of the coverage-based utility for the four test datasets made available in the supplementary files to \citet{Bansal2018}.  To account for the variability of greedy search algorithms due to initial conditions, searches are performed following each strategy for 1000 random samples of the test data of size n=1000, with a budget of B=100 queries.  We believe the superior performance of the most uncertain search exposes issues with the coverage-based utility model.  

An issue with the coverage-based utility model is that it rewards the discovery of UUs near points for which the classifier has high confidence, not the discovery of high confidence UUs themselves.  Therefore, the utility model may reward the discovery of low confidence mistakes more than the discovery of high confidence mistakes; the stated goal of the search. This is because there is no guarantee that points for which the classifier is similarly confident are confined to the same area of the feature space.  Meaning, it may be better to discover the easily found low confidence mistakes than the difficult to find high confidence mistakes. This is demonstrated by results shown in Figure~\ref{fig:coverutil}.  Upon closer inspection of the query sets, we found that the most uncertain search (unsurprisingly) discovers a larger number of UUs than the coverage-based search; indicating coverage-based utility places high value on the quantity of UUs, but will not necessarily find the UUs that provide the richest information about the classifier.

Given these apparent issues, we aim to construct a utility-based query algorithm that more appropriately rewards the identification of UUs, and helps to identify overconfident points. Again, we believe discovering query sets where UUs exist at higher rates than expected is more valuable to a rational actor than simply finding UUs.  A certain number of UUs should be expected at different confidence levels, and should be planned for.  Discovering where confidence levels are incorrect can allow better mitigation strategies. 

\section{Methodology}

We propose an alternative utility model based on facility location optimization methods  \citep{farahani2009facility}. In the facility locations problem a utility can be constructed that uses a greedy algorithm to minimize the cost, or maximize the reward, of building a series of new facilities in a supply chain, while also minimizing distances between clients to the nearest facility  [\citealt{guha1999greedy}; \citealt{arya2004local}]. In the UU query setting, we can draw an analog to the selection of a point to query to the establishment of a facility at that location in the feature space; evaluating the reward for selecting the point, and the distance it stands from the surrounding unobserved points. We propose a facility locations utility function as: 

$$W(Q) = \sum_{q \in S} r \left(c_q\right) - \frac{1}{n} \sum_{x \in \mathbb{X}} \min_{q \in S}\left(d\left(x,q\right)\right)$$

where $r\left(c_q\right) = \log($1/$(1-c_q))$ is the \textit{reward} function for finding an UU with confidence $c_{q}$, and $d(x,q)$ is the Euclidean distance between points $x$ and $q$. We use the greedy algorithm that at each iteration selects $q'$ with the maximum expected utility, as defined in Algorithm 1. 

\begin{algorithm}
	\caption{Greedy Facility Location Search}
	\label{alg:Greedy}
	\begin{algorithmic}
		\STATE {\bfseries Input:} Test set $\mathbb{X}$, prior $\hat{\phi}\left(x|Q=\emptyset\right)$, budget B
		\STATE $Q=\{\}$ \{inputs that have been queried\}
		\STATE $y_Q = \{\}$ \{oracle defined labels\}
		\STATE{\bfseries For: } $b = 1, 2, ..., B$ {\bfseries do:}

		\STATE $q' = \argmax_{q' \not\in Q} E \left[W\left(Q \cup q'\right) \right]$

		\STATE $y_{q'} = OracleQuery(q')$

		\STATE $Q \leftarrow Q \cup q'$
		\STATE $y_Q \leftarrow y_Q \cup y_{q'}$
		\STATE $S \leftarrow \left\{x | x \in Q \space \text{ and } y_x \neq M(x) \right\}$

		\STATE $b \leftarrow b + 1$

		\STATE {\bfseries Return: $Q$, $S$ and $y_Q$}
	\end{algorithmic}
\end{algorithm}

At each iterative step in Algorithm 1, we need to select the point that will maximize the expected gain in facility location utility, given probability estimates for point misclassification, $\hat{\phi}(q' | Q) = \hat{P}(y_{q'} \ne M(q' ) |  Q )$. To find the expected gain in utility for each point, we evaluate the utility under the possibilities that a point is either misclassified or correctly classified. These possible utility outcomes are then averaged with weights equal to the estimated probability of each outcome. Thus the optimization step requires the solution of the following:

$$\underset{q' \notin Q}{argmax} E[W(Q \cup q')] = $$
\begin{equation*}
\small
\underset{q' \notin Q}{argmax} \left[\begin{split}
\hat{\phi}(q') \cdot \left[\sum_{q \in S \cup q'} r \left(c_q\right) - \frac{1}{n} \sum_{x \in \mathbb{X}} \min_{q \in S \cup q'}\left(d\left(x,q\right)\right) \right] + \\ 
(1-\hat{\phi}(q')) \cdot \left[\sum_{q \in S} r \left(c_q\right) - \frac{1}{n} \sum_{x \in \mathbb{X}} \min_{q \in S}\left(d\left(x,q\right)\right)\right]  
\end{split}\right]
\end{equation*}
\normalsize

Note that $\left[\sum_{q \in S} r \left(c_q\right) - \frac{1}{n} \sum_{x \in \mathbb{X}} \min_{q \in S}\left(d\left(x,q\right)\right)\right]$ is constant for all considered points, but cannot be removed from the argmax solution because it is multiplied by an estimated probability that is unique to each point. 

In addition to a change in the utility structure from previous methods, we propose the use of model-based estimates for $\phi(x) = P\left(y_x \neq M(x) |Q \right)$. \citet{Lakkaraju2016} and \citet{Bansal2018} use different methods to provide the  estimate, $\hat{\phi}(x)$, but both are based on tracking the rate of UUs in clusters found through a multi-stage clustering procedure. The goal of this tracking is to incorporate new information into the estimation of the probabilistic structure of the search space after each step of the search algorithm. This estimation can alternatively be accomplished using model-based estimates, fit using the features and confidence scores of the query set to predict the chances of a misclassification. The posterior probability estimates that are common to most standard classification models provide a wide variety of options for creating the $\hat{\phi}(x)$ estimates. Without loss of generality, we demonstrate the use of logistic regression classifier probabilities, fitted such that:
$$\hat{\phi}(x) = logistic(c_x\hat{\beta}_0 + \sum_{j=1}^p x_j\hat{\beta}_j) = \frac{e^{c_x\hat{\beta}_0 + \sum_j x_j\hat{\beta}_j}}{1+e^{c_x\hat{\beta}_0 + \sum_j x_j\hat{\beta}_j}}$$
We select the logistic regression to demonstrate flexibility in estimating $\hat{\phi}(x)$ for several reasons: it is a generally familiar method, probabilities are the inherent model outputs, and it is computationally efficient to refit in the iterative query process. Given that fitting the logistic regression model requires at least one misclassified and one correctly classified point, we initialize the process using $\hat{\phi}(x) =(1-c_x)$ until both outcomes have been observed by the oracle.

There are a few characteristics to note in the design of the facility locations utility model. 
\begin{enumerate}
	\item In the utility function, reward is only accumulated by finding UUs in the query set.  This avoids the issue of placing value on points in the test set for simply having high confidence and being near a discovered UU. 
	\item The utility function encourages the discovery of well spread UUs by having a penalization term equal to the average minimum distance between all test points and their closest observed UU. This places value on having strong coverage of the test data by the query set, especially early in the query sequence. 
	\item The reward function, $r\left(c_x\right) = \log($1/$(1-c_x))$, is designed to impact the utility in a way that is consistent with a limiting factor being the oracle queries budget.  Further discussion of the reward function is provided below. 

\end{enumerate}

Viewed as a geometric distribution problem with a probability $\phi(x)$ of discovering a UU, we expect to need 1/$\phi(x)$ queried points like point $x$ before discovering the first UU \citep{casella2002statistical}.  For heuristic insight into the reward behavior construction, if we assume that   $\phi(x) ~= (1-c_x )$, then our reward is a log-scaled count of the number of randomly selected points we would expect to query in order to find the UUs in our query set. We use the log scaling to avoid over-incentivizing the search for incredibly rare UUs, as we know there is a limited budget for oracle queries. The optimization step will provide the highest expected rewards for selecting the most overconfident points relative to the updated probability estimates, that is to say when $(1-c_x )< \hat{\phi}(x)$. Note that unlike the UU definition, this construction does not require the arbitrary definition of a confidence threshold, $\tau$, beyond which we search for misclassifications. The reward component of the facility locations utility encourages the search procedure to select points where the model is most overconfident. We define \textit{overconfidence} as the difference between the confidence values given by the classifier and the actual rates of correct classification.

\section{Results}

We empirically evaluate our facility location utility model by applying Algorithm 1 to the four datasets used in \cite{Lakkaraju2016} and \cite{Bansal2018}: Pang04, Pang05, McAuley15 and Kaggle13.  For each dataset we fit a classifier, $M(x)$, to a biased training set, then generate predicted classes and confidence values for all observations in the test set.  We search for UUs in the test set belonging to a critical class using a feature space derived through singular value decomposition.  The datasets and classifiers were chosen to maintain consistency with the data used to evaluate both of the previous methods, unless otherwise noted.  Each dataset was obtained from the shared repository accompanying the work of \cite{Bansal2018}.  A summary for each is given below.    

\begin{itemize}
\item Pang04 \citep{pang2004}: The classification task is to label sentences from IMDb summaries and Rotten Tomatoes reviews as objective or subjective.  The dataset contains 10k sentences and the original 5k sentence training dataset was biased against objective sentences (the critical class). Bias was introduced by removing data associated with a randomly chosen leaf of a decision tree that was majority critical.  
\item Pang05 \citep{pang2005}: The classification task is to label sentences from Rotten Tomatoes reviews as positive or negative sentiment.  The dataset contains 10k sentences and the original 5k sentence training dataset was biased against negative sentences (the critical class). Bias was introduced with the method used for Pang04.  
\item McAuley15 \citep{mcauley2015}: The classification task is to label Amazon reviews as positive or negative sentiment (the critical class). The training dataset is biased because it contains 50k electronics reviews and the test set contains 5k book reviews.
\item Kaggle13 \citep{kaggle2013}: The classification task is to label images of cats and dogs.  The original 12.5k image training dataset was biased to not include black cats (the critical class) through crowd sourcing. The 5k test dataset did contain black cats. A more detailed description of the dataset creation can be found in \cite{Bansal2018}.
\end{itemize}

The classifiers for Pang04, Pang05, and McAuley15 use logistic regression with unigram features. The derived feature space used for the UUs search is created with singular value decomposition on unigram features from only the test set. The classifier for the Kaggle13 dataset is a CNN (eight convolutional layers and two linear layers), and the derived feature space is created with singular value decomposition on raw pixel values.

\begin{figure}[hbtp]
  \includegraphics[width=.45\textwidth]{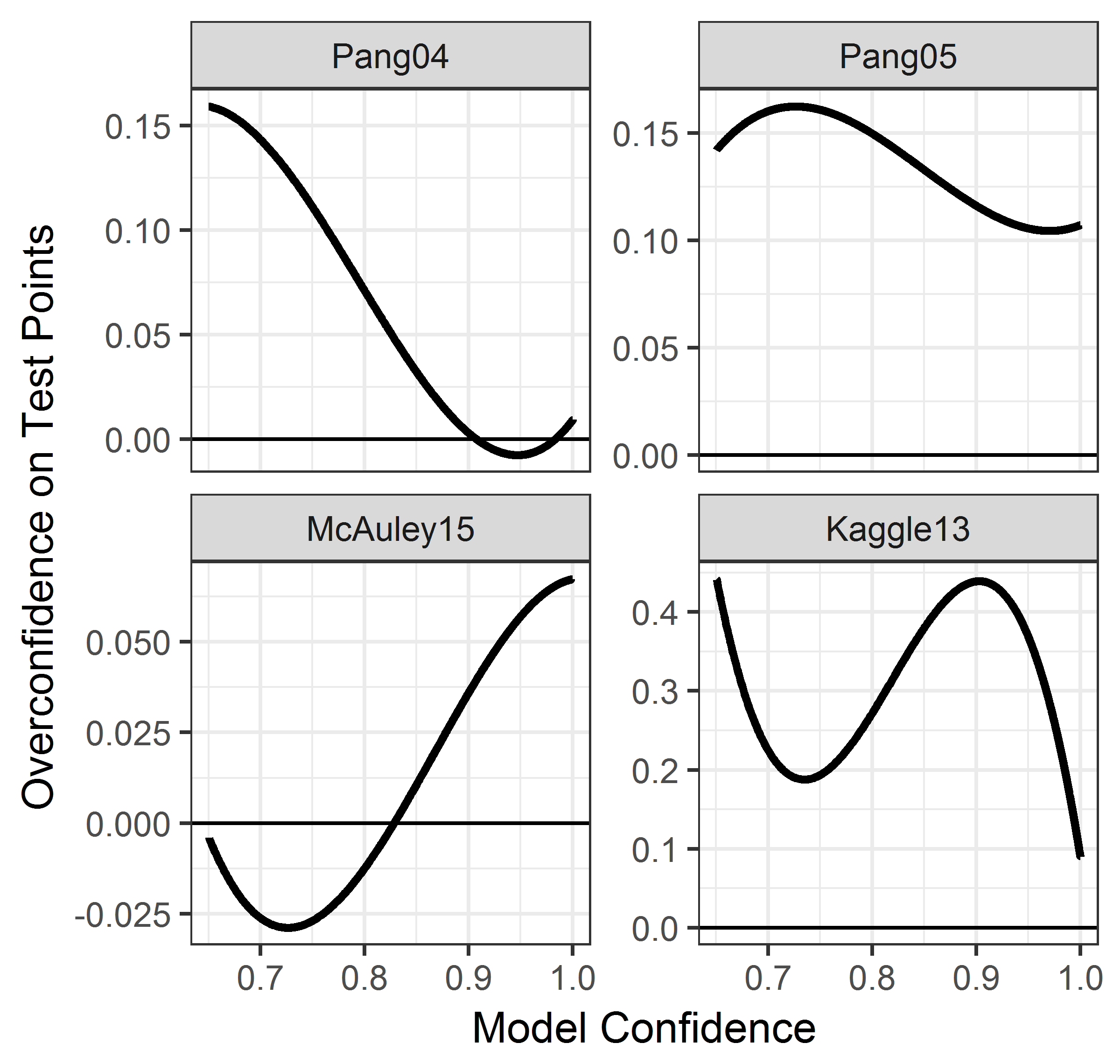}
  \caption{Observed classifier overconfidence profiles of each experimental dataset.}
  \label{fig:overconf}
\end{figure}

Figure~\ref{fig:overconf} displays the overconfidence of the models using a cubic-splines fitted between the indicator of correct classification and the confidence value in each test dataset to obtain an empirical estimate of the true rates of correct classification. We see that the models from the Pang04 and Pang05 datasets are most overconfident for points with relatively low confidence values, thus we would expect that a simple sequential search of the most-uncertain points to provide high facility locations utility. The predictions for McCauley15 and Kaggle13 are most overconfident for points in the higher confidence range, thus most-uncertain search should provide low facility locations utility. We see that these four datasets represent fairly different profiles of overconfidence, thus present good variety for evaluating characteristics of the facility locations utility model. 

As with the evaluation of the coverage-based utility, we run the facility locations queries on 1000 random samples of size n=1000 from each of the datasets, using a budget B=100. In the following subsections we evaluate the utility outcomes of the facility locations queries in comparison to the most-uncertain search method, and compare the ability of several algorithms to discover overconfident points.

\subsection{Facility Location Utility Outcomes}

To evaluate the queries generated by the facility locations utility model we collect query results from running Algorithm~\ref{alg:Greedy} on repeated random samples from the test sets, thus allowing Monte Carlo estimates to be used for utility characteristics. 

\begin{figure*}[t]
 \centering
  \includegraphics[width=\textwidth]{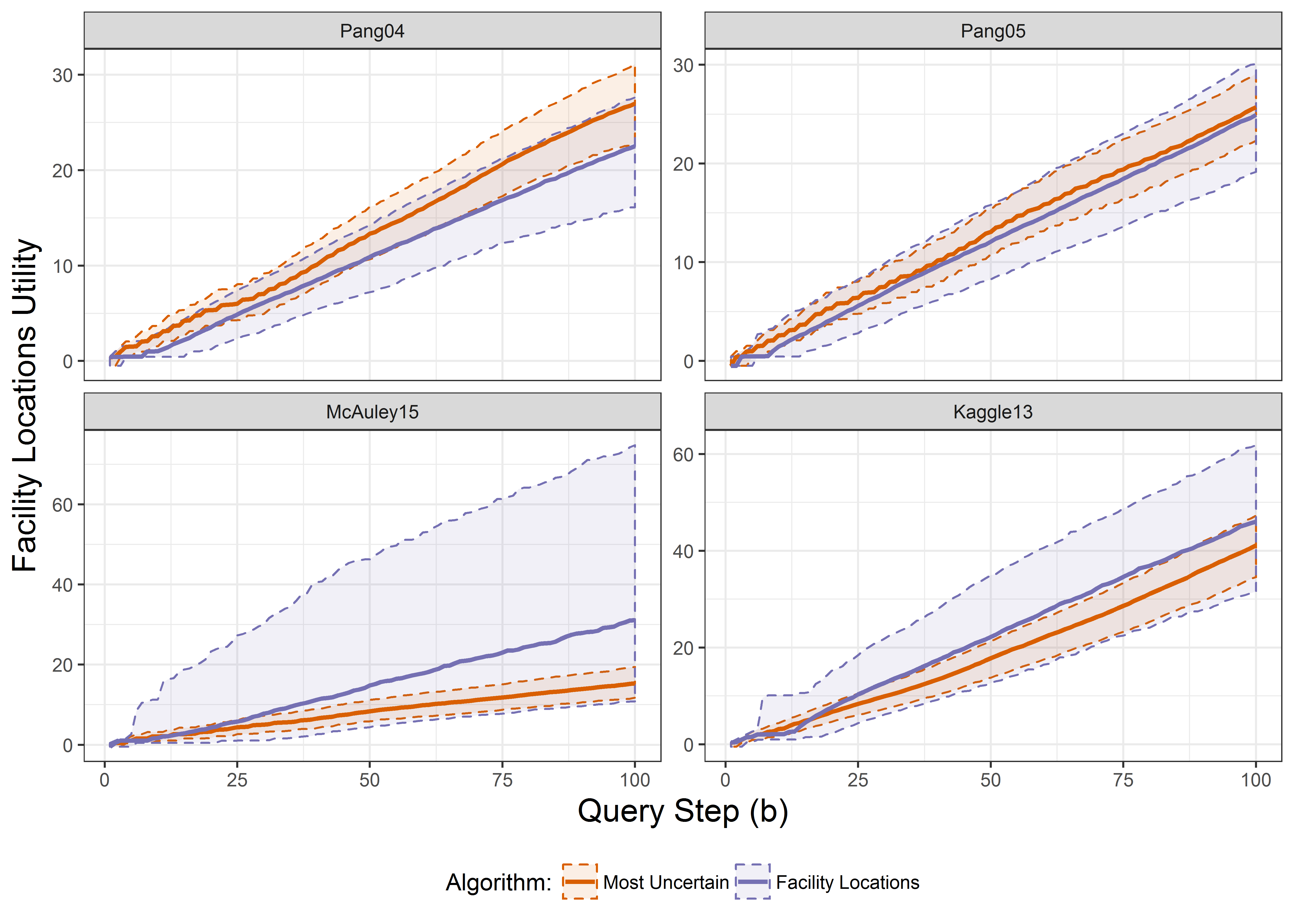}
  \caption{Comparison of facility locations utility outcomes achieved under most uncertain and coverage-based search algorithms. Monte Carlo medians (solid) and 90\% prediction bands (dashed).  Higher facility locations utility is typically achieved with facility locations search in all cases.}
  \label{fig:flutil}
\end{figure*}

Figure~\ref{fig:flutil} displays the Monte Carlo medians and 90\% predictions bands for the facility locations utility gains. We see that the most-uncertain selection method --- that begins its search with points with confidence values just above $\tau$=.65 --- provides the strongest utility for the Pang04 case, as is desired, because the overconfidence profile in Figure~\ref{fig:overconf} shows the highest overconfidence for points in this range. The utility values from facility locations search were typically slightly lower. For the Pang05 case where the overconfidence profile is skewed right but relatively ubiquitous, we find comparable utility outcomes between the facility location search and the most-uncertain search. For the McAuley15 where overconfidence profile is heavily skewed left, we see inconsistent, but higher utility outcomes from the facility locations search, and consistently low utility outcomes from the most-uncertain search. In the last case of Kaggle13, where the overconfidence profile is multi-modal, the facility locations search provided less consistent, but typically stronger utility outcomes than most-uncertain searches. Thus in all scenarios, the facility locations utility model is properly placing value on the pursuit of the most overconfident points, as per its design.

\subsection{Efficient Discovery of Overconfidence}

We compare the queries gathered by the coverage-based utility algorithm from \cite{Bansal2018}, the bandit search algorithm from \cite{Lakkaraju2016}, and our facility locations utility algorithm. Given that all of the searches rely on their own utility function, it does not make sense to compare their selections on the utility values directly. Instead we compare the efficiency of the search for UUs, using a summary statistic that we call the \textit{standardized discovery ratio} (SDR). The SDR is an adaptation of the \textit{standardized mortality ratio} used in biostatistics to evaluate the mortality rate for a given sample of patients, which standardizes using their initial risk of death [\citealt{taylor2013standardized}; \citealt{rosner2015fundamentals}]. In our case we use an analog that evaluates the misclassification rate, standardized by the initial model confidence. The SDR is computed as 
$$ |S| / \sum_{x=1}^B(1- c_x) $$
thus counting the number of discovered misclassifications, divided by the number of misclassifications expected based on the confidence values of the queried points. The SDR can be interpreted as the number of times more misclassifications were found than were expected based on model confidence; making it a natural metric for evaluating overconfidence.

\begin{figure}[hbtp]
  \includegraphics[width=.49\textwidth]{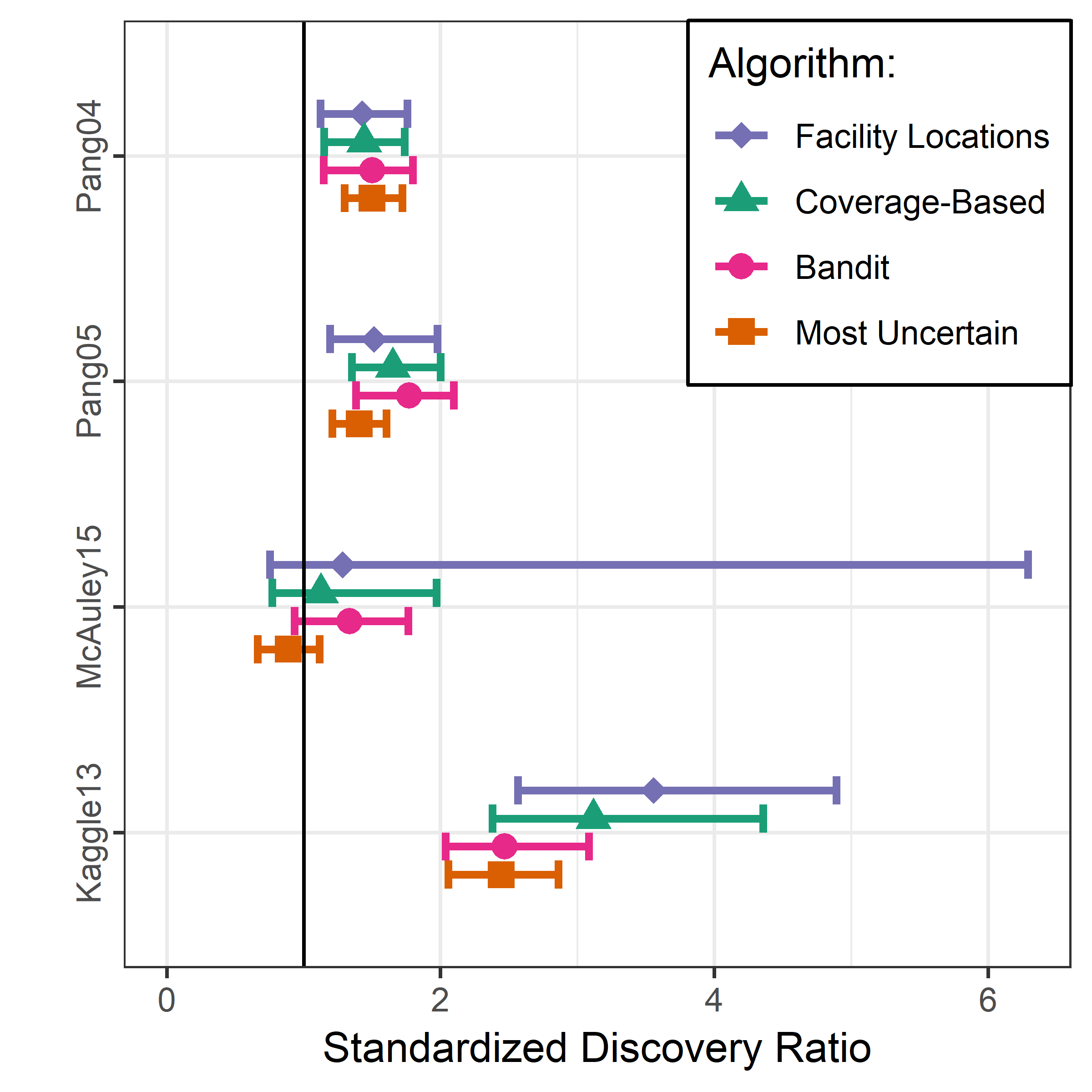}
  \caption{Monte Carlo medians and 90\% prediction intervals of the standardize discovery ratio.}
  \label{fig:sdr}
\end{figure}

Figure~\ref{fig:sdr} compares the Monte Carlo medians and 90\%  central prediction intervals for the SDR values associated with 1000 random samples of size n=1000 from each of the datasets, using each of the four query algorithms: facility locations, coverage-based, bandit, and most uncertain. The SDR intervals for Pang04 and Pang05 reveal that all four algorithms are similarly efficient at discovering overconfident UUs in situations where the overconfident points fall just beyond the defined threshold, $\tau = 0.65$. The SDR intervals for McAuley15 and Kaggle13, where overconfidence was most prevalent for points far beyond the threshold, the facility locations utility algorithm typically provides the most efficient discovery of overconfident points. For Kaggle13, the median SDR for the facility locations algorithm is 1.2 times larger than the coverage-based utility algorithm and 1.6 times larger than both the most uncertain and bandit algorithms.

\section{Discussion \& Conclusions}

Previous literature has defined unknown unknowns as any highly confident predictions that result in misclassification, possibly with respect to a critical class. This definition ignores the unavoidable uncertainties of predictive modeling. It should be expected that classifier predictions are imperfect, this is why confidence values exist! The actions taken as a result of the predictions should take into account the inherent uncertainty. However, in the case where the claimed confidence is overstated, a rational actor cannot properly mitigate the risk posed by misclassification. Unlike the previous works that propose utility functions that seek to uncover high confidence misclassifications, the facility locations utility that we propose is designed to seek out \textit{overconfident} misclassifications. 

Through repeated random initialization in our computational experiments, we thoroughly tested the outcomes of our facility locations utility algorithm against the bandit, coverage-based, and most uncertain search algorithms. We have demonstrated the ability of our greedy algorithm, using logistic regression probability estimates for $\hat{\phi}(x)$ in the optimization step, to consistently obtain strong facility locations utility in four data scenarios with disparate overconfidence profiles. This is important because in real-world applications we would not know the overconfidence behavior a priori to our query search, so we require a versatile estimation method. We have also demonstrated that oracle queries gathered using a facility locations utility search tend to have higher standardized discovery ratios than the alternative algorithms, thus represent a more efficient use of the constrained budget for queries. 

The source code and datasets needed for replicating the experimental results discussed in this paper are available online in the supplemental materials for this manuscript. Also, an on open source  implementation in R \citep{R} of the facility locations algorithm and associated functions through the \km{\texttt{uuutils}} R package can be accessed through the github repository at \km{\url{www.github.com/kmaurer/uuutils}}.

There are many avenues for future work related to the facility locations utility methods that we have presented. First, the facility locations utility model structure separates the discovery reward and coverage proximity components, which could allow separate rescaling to weight each component in line with the priorities of an application, or use of non-Euclidean distances. Next, exploratory methods could be developed to evaluate what the query set tells us about the overconfidence of your model, perhaps interpreting the structure of the models used to predict $\hat{\phi}(x)$ to better understand what features are related to overconfidence. Additionally, we have proposed a greedy solution for the facilities location problem that updates after each oracle query, but non-sequential solutions could be employed to select sets of points to query at each iteration. Lastly, there may be cases where it is impractical to collect a large enough oracle query set to refit the original classifier, but it may be sufficient to estimate the original classifier’s overconfidence and perform recalibration so that actions taken based on the predictions can include more appropriate risk mitigation. 

\section{Acknowledgments} 

\km{Support for this research was provided by the Visiting Summer Faculty Program and Extension Grants sponsored by the Griffiss Institute, in collaboration with the Air Force Research Lab. 

We would like to thank Dr. Eric Heim, Dylan Elliott, and Dr. John Bailer for providing valuable feedback throughout the development of these methods. 

Put Air Forces - "cleared for public release statement" here}

\newpage
\bibliographystyle{aaai}

\bibliography{librarySmall}

\end{document}